\documentclass[10pt,twocolumn,twoside]{IEEEtran}
\usepackage{amsmath,amssymb,amsfonts,amsthm,mathtools}
\usepackage{subfigure}

\ifCLASSINFOpdf
\else
\fi
\newcommand{\R}{\mathcal{R}}

\newcommand{\F}{\mathcal{F}}

\newcommand{\U}{\mathcal{U}}

\newcommand{\1}{\textbf{1}}
\usepackage{latexsym}

\usepackage
	{todonotes}

\begin{document}

\title{Whittle index based Q-learning for restless bandits with average reward\thanks{Work is supported in part by
the DST-Inria project ``Machine Learning for Network Analytics'' IFC/DST-Inria-2016-01/448.
VB is also supported in part by a J.\ C.\ Bose Fellowship from the Government of India and
KA is also supported in part by Nokia Bell Labs and ANSWER project PIA FSN2
(P15 9564-266178 \textbackslash  DOS0060094). }}


\author{
\IEEEauthorblockN{Konstantin E. Avrachenkov\IEEEauthorrefmark{1}, Vivek S.\ Borkar\IEEEauthorrefmark{2}}\\
\vspace{0.05in}
\IEEEauthorblockA{\IEEEauthorrefmark{1}Inria Sophia Antipolis, 2004 Route des Lucioles, Valbonne 06902, France,
\emph{k.avrachenkov@inria.fr}}\\
\IEEEauthorblockA{\IEEEauthorrefmark{2}Indian Institute of Technology, Powai, Mumbai, 400076, India, \emph{borkar.vs@gmail.com}}
\vspace{0.05in}
}

\maketitle

\begin{abstract}
A novel reinforcement learning algorithm is introduced for multiarmed restless bandits with average reward, using the paradigms of Q-learning and Whittle index. Specifically, we leverage the structure of the Whittle index policy to reduce the search space of Q-learning, resulting in major computational gains. Rigorous convergence analysis is provided, supported by numerical experiments. The numerical experiments show excellent empirical performance
of the proposed scheme.
\end{abstract}

\IEEEpeerreviewmaketitle



\section{Introduction}

Restless bandits have found numerous applications for various scheduling and resource allocation problems, such as wireless communication \cite{Aalto,Liu,Raghu}, web crawling \cite{Allerton,ephemeral,Nino14},
congestion control \cite{Aetal13,ABSCDC,Aetal18},
queueing systems \cite{Glaze4,proc_sharing,Glaze3,Larranaga}, cluster and cloud computing \cite{Cloud,Nino2},  machine maintenance \cite{Glaze05}, target tracking \cite{NinoVillar} and clinical trials \cite{Villaretal2015}. See \cite{Gittins,Jacko,Ruiz} for book-length accounts of theory and applications of restless bandits. While restless bandits can be viewed as a special case of classical Markov decision processes, this suffers from curse of dimensionality because the state space grows exponentially in the number of arms. In fact, the problem is provably hard in the sense of belonging to the complexity class PSPACE \cite{Papa}. One very successful heuristic in this context has been the celebrated Whittle index policy \cite{Whittle}, which firstly relaxes the `hard' constraint of using a certain number of arms at each time, to doing so on the average. Thereby it allows a decoupling of the problem into multiple individual controlled Markov chains via the Lagrange multiplier, using the fact that both the reward and the constrained functional are separable. This leads to a state space that grows linearly in the number of arms. Then, secondly, these chains are coupled through the control policy based on ordinal comparison of a scalar function of their individual states, viz., the so-called Whittle index. While this is known not to be optimal in general, it works very well in practice and is asymptotically optimal in a certain sense \cite{Larranaga,Weber}.

The use of Whittle index policy, however, requires full knowledge of the system, both in the relatively few cases where it is known explicitly (e.g., in \cite{ephemeral}) or when it has to be numerically calculated. This is often not the case in practice, sometimes called the `curse of modeling'. The uncertainties can be either parametric or structural, or both. In either case,  the classical adaptive control schemes (e.g., \cite{Hernandez-Lerma}) or off-the-shelf reinforcement learning schemes (e.g., \cite{Bertsekas}, \cite{NeuroDP},  \cite{SuttonBarto}) become computationally unmanageable if applied directly to restless bandits. These schemes typically do not exploit the special structure available in the problem, in this case the Whittle indexability.

In this work, we combine the  Q-learning algorithm for average reward \cite{Abounadi} with a tuning scheme for the Whittle indices. This yields a provably convergent learning algorithm with excellent empirical performance on test cases. In case the arms are statistically identical, the algorithm is particularly economical because it learns the common Q-values and Whittle index. The algorithm has a notably simple form compared to above works, and can be executed in both on-line and off-line modes, the latter allowing for off-policy iterations.

The main novelty of our work lies in the fact that our Q-learning algorithm is tuned to a specific  policy class, viz., the Whittle index policy. This drastically reduces the search space for policy by exploiting a specific additional structure in the problem, viz., its Whittle indexability. The task is now to learn the Whittle indices. If $d_1, \cdots, d_N$ are the cardinalities of the state spaces of the $N$ bandit's arms, then our scheme needs $\sum_{\alpha=1}^N(2d_\alpha^2 + d_\alpha)$ updates per iteration, whereas a vanilla Q-learning would need $2^N\prod_{\alpha=1}^N d_\alpha$ updates. For $d_\alpha \equiv 5 \ \forall i$ and $N = 100$, this number is $10^{100}$ which already exceeds the number of atoms in the universe. Compare this with our scheme where the corresponding number is $5500$ and the scheme is efficient on most standard computational devices.  In fact this issue persists even after using function approximation, e.g., DQN \cite{Mnih2015}, because the maximization operator requires searching a large action space. We elaborate on this in Section~4.
On the other hand, compared to earlier works along similar lines, we have the advantage that our scheme does not assume additional structure such as an optimal threshold policy for decoupled chains \cite{Chadha} or explicit formula for the Whittle index \cite{Allerton}. In addition, \cite{Chadha} lacks a completely rigorous convergence proof whereas \cite{Allerton} has numerical issues and is useful only when the model is known but its parameters are not. Another precursor is \cite{Moka} which does not have a rigorous proof,  systematically underestimates the reward, and converges to a random limit.

%
%
%

 A related line of work \cite{Roy1}, \cite{Roy2} develops learning schemes for threshold policies. The Whittle index itself, however, is not a simple threshold, but a function of the state. So the problems are quite distinct.   Furthermore, \cite{Roy1}, \cite{Roy2} use policy gradient based methods  as in \cite{Marbach}  which are better suited when a  low dimensional parameter such as a threshold is involved. In contrast, Whittle index is a function on the state space for which they are ill suited. At the same time, Whittle index is defined in terms of an equality. So a much simpler scheme is used here, which makes incremental changes towards forcing this equality.

 It is worth noting that both the present work as well as some others such as primal-dual type schemes for constrained Markov decision processes \cite{BorkarConstrained} fall within the ambit of the larger paradigm of using two time scale iterations when both control and parametric optimization are simultaneously present, and it makes sense to perform the latter on a slower time scale to reduce the problem to a bi-level optimization, which makes it computationally more tractable.

The paper is organized as follows. The next section summarizes  the Whittle index formalism. Section 3 describes the algorithm in detail. Section 4 presents numerical experiments. Section 5 provides  convergence analysis, which relies upon \cite{Abounadi}, \cite{Bhat0}, \cite{BorkarBook}, \cite{Karmakar} and \cite{Bhat}.

The notation will be as follows. The subscripts $n,m$, resp., $t$, will stand for discrete and continuous time indices. For the Q-learning iterates, we use superscript $\alpha$ to indicate the $\alpha$-th arm. Another superscript $(c)$
denotes a scaling factor. Letters $i, j, k$ etc.\ indicate elements of the state space. When we want to keep a state variable fixed during an iteration, we distinguish it as $\hat{i}, i^*$ etc. We use letters $u, v$ to denote the control variables. We use $[[q_{ij}]]$ to denote a matrix whose $(i,j)$th entry is $q_{ij}$. Letter $\lambda$ will denote the Whittle subsidy, also used as a subscript when required.

\section{Whittle index for restless bandits}

We first recall briefly the Whittle paradigm \cite{Whittle}. The restless bandit problem consists of a finite collection of (say) $N$ Markov chains, each with two possible modes of evolution, active and passive. These correspond to two possibly distinct transition matrices and reward functions that may depend on the particular chain under consideration. The problem is to keep active exactly (or at most) $M < N$ out of $N$  chains so as to maximize the reward. The passive chains also evolve, albeit with a different transition matrix, instead of remaining frozen in the state they occupied at the time of turning passive. This makes this scenario distinct from the classical multiarmed rested bandits. The latter problem has an explicit optimal solution in terms of the Gittins index policy \cite{Gittins}, which assigns to each chain a function of its state called the Gittins index. At every time instant, given the current state profile, the corresponding indices are sorted in a decreasing order and the top $M$ rendered active, ties being resolved arbitrarily. The restless bandit problem, however, is much harder, provably so as already noted. Whittle replaced it by a more tractable relaxation wherein the per stage constraint of $M$ out of $N$ chains is relaxed to an average constraint that requires only the asymptotic fraction of active chains to be $M$ (this is made precise below). This alone is not enough. Motivated by the classical Lagrange multiplier formulation for this problem, Whittle introduced a `subsidy' for passivity and defined the problem to be (Whittle) indexable if the set of passive states increases monotonically from the empty set to the whole state space when the subsidy is increased from $-\infty$ to $\infty$. In this case, he defined the (Whittle) index to be the value of the subsidy, as a function of the current state, for which both active and passive modes are equally desirable. The Whittle policy then is to sort these indices in a decreasing order for the current state profile of the chains, and render the top $M$ active. While not necessarily optimal, the policy is known to do well in practice and is provably optimal in a certain limiting sense as mentioned above\footnote{Note that this work is in the domain of `Markov bandits', a branch of Markov decision processes, distinct from the current activity on multiarmed bandits in machine learning that deals with rewards independent across arms and time, or their dependent variants quite distinct from Markov bandits \cite{Lattimore}. Their objective is also different, viz., to bound the asymptotic regret, unlike the Markov bandits which seek optimality for the classical criteria of Markov decision theory. Work on Markov bandits in machine learning community is only recently beginning to pick up, see, e.g., \cite{Kumar}.}.

Specifically, we consider $N > 1$ controlled Markov chains
$\{X^\alpha_n, n \geq 0\}$, $1 \leq \alpha \leq N,$
on a finite state space $S = \{1,2, \cdots , d\}$, $1 < d < \infty$ with control space $\U := \{0, 1\}$. The controlled  transition kernel
$$(i,j,u) \in S^2\times\U  \mapsto p^\alpha(j | i, u) \in [0, 1]$$
for the $\alpha$-th chain satisfies $\sum_j p^\alpha(j | i, u) = 1 \ \forall \ i, u,$ and has the interpretation of `probability of going from state $i$ to state $j$ under control $u$'. The control variable $u$ is binary, corresponding to two modes of operation, active ($u = 1$) and passive ($u = 0$).
Define the increasing family of $\sigma$-fields  $\F_n :=$ $\sigma(X_m^\alpha, U_m^\alpha, 1 \leq \alpha \leq N, m \leq n)$, $n \geq 0$. The `controlled Markov property' is
$$
P(X^\alpha_{n+1} = j | \F_n) = p^\alpha( j | X^\alpha_n, U^\alpha_n), \ \forall \ n \geq 0, \ 1 \leq \alpha \leq N,
$$
where $\{U^\alpha_n\}_{n \geq 0}, 1 \leq \alpha \leq N,$ are the $\U $-valued control processes, called `admissible controls'. A special subclass denoted SP is that of stationary policies wherein $U^\alpha_n = \varphi^\alpha(X^\alpha_n)$ for some $\varphi^\alpha : S \mapsto \{0,1\}$, $n \geq 0$.  The individual chains are called `arms' of the restless bandit. Let $r^\alpha : (i,u) \in S\times\U  \mapsto \R$ denote prescribed per stage reward function for the $\alpha$-th chain. These controlled Markov chains are assumed to satisfy:

\noindent \textbf{(C0)} (Unichain property) There exists a distinguished state $i_0 \in S$ that is reachable with strictly positive probability from any other state under any stationary policy (SP).

 Since $d = |S| < \infty$, this implies in particular that for $\tau := \min\{n \geq 0 : X^\alpha_n= i_0\}$,
    \begin{equation}
    \max_{k \in S, SP}E[\tau | X^\alpha_0 = k] < \infty. \label{hittingtime}
    \end{equation}
The objective is to maximize the long run average reward
\begin{equation}
\liminf_{n\uparrow\infty}\frac{1}{n}E\left[\sum_{m=0}^{n-1}\sum_{\alpha=1}^Nr^\alpha(X^\alpha_m, U^\alpha_m)\right], \label{reward}
\end{equation}
subject to the constraint: for a prescribed $M < N$,
\begin{equation}
\sum_{\alpha = 1}^NU^\alpha_n = M, \quad \forall n. \label{constraint}
\end{equation}
That is, at each time instant, only $M$ arms are activated.

The Whittle relaxation is to replace the `per time instant' constraint (\ref{constraint}) by a `time-averaged constraint'
\begin{equation}
\liminf_{n\uparrow\infty}\frac{1}{n}E\left[\sum_{m=0}^{n-1}\sum_{\alpha=1}^NU^\alpha_m\right] = M. \label{constraint2}
\end{equation}
This renders it a classical `constrained Markov decision process' \cite{Altman}. While this is a significant simplification, the problem is still unwieldy. Whittle's ingenious observation was to use the fact that it is a problem with separable cost and constraint and invoke the Lagrangian relaxation to decouple it into individual control problems given the Lagrange multiplier $\lambda$.  That is, we consider now the unconstrained control problem of maximizing
 \begin{equation}
\liminf_{n\uparrow\infty}\frac{1}{n}E\left[\sum_{m=0}^{n-1}(r^\alpha(X^\alpha_m, U^\alpha_m) + \lambda(1 - U_m^\alpha)) \right] \label{reward}
\end{equation}
separately for each $\alpha$.\footnote{We have dropped a constant factor involving  $M$  from the total reward so as to match it with Whittle's  set-up. This does not affect the optimization problem because we finally do an ordinal comparison that is unaffected by this.} The dynamic programming equation then is
\begin{eqnarray}
\lefteqn{V^\alpha(i) = \max\Big(r^\alpha(i,1) + \sum_j p^\alpha(j | i, 1)V^\alpha(j),} \nonumber \\
&& r^\alpha(i,0) +  \lambda + \sum_j p^\alpha(j | i, 0)V^\alpha(j)\Big) - \beta^\alpha \label{DP}\\
&=& \max_{u \in \U}\Big(u(r^\alpha(i,1) + \sum_j p^\alpha(j | i, 1)V^\alpha(j)) + (1-u)\times \nonumber \\
&& (r^\alpha(i,0) +  \lambda + \sum_j p^\alpha(j | i, 0)V^\alpha(j))\Big) - \beta^\alpha, \label{DP2}
\end{eqnarray}
with $(V^\alpha(\cdot), \beta^\alpha) \in \R^d\times\R$ the unknown variables. Under \textbf{(C0)}, $\beta^\alpha$ is unique and equals the optimal reward. $V$ is unique up to an additive constant.
The optimal decision $u^*(i)$ in state $i$ then is given by the maximizer in the right hand side of (\ref{DP}) \cite{Puterman}. If we consider individual arms separately, the superscript $\alpha$ will be dropped henceforth, used only when it is needed. Note, however, that this does not mean that all arms are assumed statistically identical.

Define the Q-value as
\begin{eqnarray}
Q(i,u) &:=& u(r(i,1) + \sum_j p(j | i, 1)V(j)) + (1 - u)\times \nonumber \\
&& (r(i,0) +  \lambda + \sum_j p(j | i, 0)V(j))) - \beta.
\end{eqnarray}
This satisfies the equation
\begin{eqnarray}
Q(i,u) &=& ur(i,1) + (1-u)(\lambda + r(i,0)) - \nonumber \\
&&\beta + \sum_j p(j|i,u)\max_v Q(j,v), \label{Q-DP}
\end{eqnarray}
for $i \in S, u \in U$. Under \textbf{(C0)}, this has a  solution $(Q, \beta)$ where $\beta$ is uniquely specified as the optimal reward and $Q$ is unique up to an additive scalar, just as for (\ref{DP}). The set $\{j \in S : u^*(j) = 1\}$ is the set of states when the arm is active, its complement being the set of states when it is passive. Whittle's insight was to view the Lagrange multiplier as a `subsidy' for passivity. He  defined the problem to be indexable when the set of passive states increases monotonically from the empty set to all of $S$ as the subsidy is increased from $-\infty$ to $\infty$. In this case, he defines the (Whittle) index for state $\hat{k}$ to be the value $\lambda(\hat{k})$ of $\lambda$ for which both active and passive modes are equally preferred in state $\hat{k}$. That is,
\begin{eqnarray}
\lambda(\hat{k}) &:=& r(\hat{k},1) + \sum_jp(j | \hat{k}, 1)V(j)
- r(\hat{k},0) \nonumber \\
&& \ \ \ \ \ \ \ \ \ - \  \sum_jp(j | \hat{k}, 0)V(j). \label{Windex0}
\end{eqnarray}
This is equivalent to solving
\begin{equation}
Q(\hat{k}, 1) - Q(\hat{k}, 0)=0,  \label{Windex}
\end{equation}
for $\lambda = \lambda(\hat{k})$, where the $\lambda$-dependence of the left hand side is not rendered explicit as per our convention thus far.

Our algorithm is a two time scale iteration wherein the faster timescale performs Q-learning for a `static' subsidy $\lambda_n$, the latter in reality changing on a slower time scale. Thus it tracks the Q-value corresponding to the slowly changing subsidy, which in turn is updated on a slower timescale by a simple tuning scheme suggested by (\ref{Windex}). The Whittle index is a function of $\hat{k} \in S$, so for large state spaces, one may compute it for a suitably chosen subset of $S$ and interpolate.

\section{Q-learning for Whittle index}

Q-learning is one of the oldest and most popular reinforcement learning scheme for approximate dynamic programming, due to Watkins \cite{Wat}. Originally developed for infinite horizon discounted rewards, we shall be using a variant for average reward from \cite{Abounadi}. For the controlled Markov chain  $\{X^\alpha_n\}$ above with average reward (\ref{reward}),  the `RVI Q-learning' algorithm of (2.7) in \cite{Abounadi} is as follows (with a key difference we highlight later).
Fix a stepsize sequence $\{a(n)\}$ satisfying $\sum_na(n) = \infty$ and $\sum_na(n)^2 < \infty$. For each $i \in S, \ u \in \U $, do:
\begin{eqnarray}
Q_{n+1}(i,u) &=& Q_n(i,u)  +   a(\nu(i,u,n))I\{X_n = i, U_n = u\} \nonumber \\
&& \times \Big((1 - u)(r(i,0) + \lambda) + ur(i,1) +  \nonumber \\
&& \max_{v\in\U}Q_n(X_{n+1},v)  - f(Q_n) - \ Q_n(i,u)\Big),  \nonumber \\
&& \  \label{Q-update}
\end{eqnarray}
where\footnote{This is not the unique choice of $f(\cdot)$, see \cite{Abounadi}. }
\begin{equation}
f(Q) = \frac{1}{2d}\sum_{i\in S}(Q(i,0) + Q(i,1)). \label{eff}
\end{equation}
Here for $i \in S, u \in \U $,
$$\nu(i,u,n) = \sum_{m=0}^nI\{X_m = i, U_m = u\},$$
is the `local clock' for the pair $(i,u)$ counting the updates of the $(i,u)$-th component.

Our objective is to learn the Whittle index, i.e., the value $\lambda(\hat{k})$ of $\lambda$ defined in (\ref{Windex0}), equivalently in (\ref{Windex}), for which active and passive modes are equally desirable for a given $\hat{k} \in S$. Hence we also have an updating scheme for $\lambda$, leading to a coupled iteration for each $\hat{k} \in S$. The first component is the same as (\ref{Q-update}) except for the replacement of $\lambda$ by the estimated Whittle index $\lambda_n(\hat{k})$. Thus for each $\hat{k} \in S$, we perform the iteration
\begin{eqnarray}
\lefteqn{Q_{n+1}(i,u;\hat{k}) = Q_n(i,u;\hat{k})  +  a(\nu(i,u,n)) \times} \nonumber \\
&&  I\{X_n = i, U_n = u\}\Big((1-u)(r(i,0) + \lambda_n(\hat{k})) +  ur(i,1)    \nonumber \\
&& + \ \max_{v\in\U}Q_n(X_{n+1},v;\hat{k})  - f(Q_n{(\hat{k})}) - \ Q_n(i,u;\hat{k})\Big) \nonumber \\
& \  \label{Q-update0}
\end{eqnarray}
along with an update for learning the Whittle index $\lambda(\hat{k})$ for state $\hat{k}$ given by:
with a prescribed stepsize sequence $\{b(n)\}$ satisfying $\sum_nb(n) = \infty$, $\sum_nb(n)^2 < \infty$ and $b(n) = o(a(n))$, do
\begin{equation}
\lambda_{n+1}(\hat{k}) = \lambda_n(\hat{k}) + b(n) \left( Q_n(\hat{k},1;\hat{k}) - Q_n(\hat{k},0;\hat{k}) \right).
\label{lambda-update}
\end{equation}
We use the `hat notation' $\hat{k}$ to emphasize that $\hat{k}$ is the $\hat{k}$-th component of the Whittle index
estimation evolving on the slow time scale.

Note that we need to estimate $Q(i,u;\hat{k})$ for each arm $\alpha$. However, if some arms are statistically
identical, we can take advantage of this and collect statistics simultaneously from statistically identical arms.
For instance, in the case of homogeneous arms and shared memory architecture, we need to update only
$2d^2+d$ variables, whereas they would have been $(2d)^N$ with Q-learning applied directly without the Whittle scheme.\\

The control actions  at time $n$ are defined as follows: Let $0 < \epsilon < 1$ be prescribed. With probability $(1 - \epsilon)$, we sort arms in the decreasing order of the estimated Whittle indices $\lambda_n(X_n^\alpha), 1 \leq \alpha \leq N,$ and render the top $M$ arms active, the remaining arms are passive. Ties  are broken according to some  pre-specified convention. With  probability $\epsilon$, we render active $M$ random arms, chosen uniformly and independently,  the rest passive.
This uniform randomization with probability $\epsilon$ is essential in order to ensure that all state-action pairs are sampled `frequently', i.e.,
\begin{equation}
\liminf_{n\uparrow\infty}\frac{\nu(i, u, n)}{n+1} \geq \Delta \ \ \mbox{a.s.} \label{frequent}
\end{equation}
for some $\Delta > 0$ and all $i, u$. This is because by the martingale law of large numbers,
\begin{eqnarray*}
\lefteqn{\lim_{n\uparrow\infty}\frac{1}{n}\Big(\sum_{m=0}^n(I\{X_{m+1} = i, U_{m+1} = u\} } \\
&& \ \ \ \ \ \ \ - \ P(U_{m+1} = u|i)p(i|X_m, U_m))\Big) = 0 \ \mbox{a.s.}
\end{eqnarray*}
and therefore the l.h.s.\ of (\ref{frequent}) equals
\begin{eqnarray*}
\lefteqn{\liminf_{n\uparrow\infty}\frac{\sum_{m=0}^nI\{X_{m+1} = i, U_{m+1} = 0\}}{n}} \\
&\geq& \ \ \frac{\sum_{m=0}^n\epsilon p(i|X_m, U_m)}{n} \\
&\geq& \ \ \epsilon\min\{p(k|j,v) : p(k|j,v) > 0, k, j \in S, v \in \U \} > 0.
\end{eqnarray*}
This ensures (\ref{frequent}), i.e., adequate exploration, though at the expense of settling for near-optimality rather than optimality. The prospect of slowly decreasing $\epsilon$ with $n$ has been explored in literature, see, e.g., \cite{Singh}.

We used following  stepsize
sequences, which gave good performance in practice:
\begin{equation}
a(n) = \frac{C}{\lceil\frac{n}{500}\rceil}, \
b(n) = \frac{C'}{1 + \lceil\frac{n\log n}{500}\rceil}I\{n (\mbox{mod} \ N) \equiv 0\}.
\label{step_sizes}
\end{equation}

Define $h(Q, \lambda) = [[h(Q, \lambda)_{iu}]]_{i \in S, u \in \{0,1\}}: \R^{2d}\times\R \mapsto \R^{2d}$
by
\begin{eqnarray}
h(Q, \lambda)_{iu} &:=& (1-u)(r(i,0) + \lambda) \ +
ur(i,1) \label{Hech} \\
&&+ \sum_jp(j|i,u)\max_{v\in\{0,1\}}Q(j,v) - f(Q). \nonumber
\end{eqnarray}
Letting $Q(\hat{k})$ denote $Q(\cdot,\cdot;\hat{k})$ suitably vectorized, we also define $M_n{(\hat{k})} := [[M_n(\hat{k})_{iu}]]_{i \in S, u \in \{0,1\}}$ by
\begin{eqnarray}
\lefteqn{M_{n+1}{(\hat{k})}_{iu} := (1-u)(r(i,0) + \lambda_n(\hat{k})) +
ur(i,1)  \ + }  \nonumber \\
&& \max_{v\in\U}Q_n(X_{n+1},v;\hat{k})  - f(Q_n(\hat{k})) - h(Q_n{(\hat{k})}, \lambda_n(\hat{k}))_{iu}\Big). \nonumber \\
&& \ \label{emm}
\end{eqnarray}
Then $\{M_n{(\hat{k})}\}$ are martingale difference sequences w.r.t.\  $\{\F_n\}$, i.e., they are adapted to $\{\F_n\}$ and satisfy $E[M_{n+1}(\hat{k})_{iu} | \F_n] = 0 \ \forall \ i,\hat{k},u,n$. Rewrite (\ref{Q-update0}) as
\begin{eqnarray}
&& Q_{n+1}(i,u;\hat{k}) = Q_n(i,u;\hat{k}) \nonumber \\
&& + \ a(\nu(i,u,n))I\{X_n = i, U_n = u\} \nonumber \\
&&\times(h(Q_n{(\hat{k})}, \lambda_n(\hat{k}))_{iu} - Q_n(i,u;\hat{k}) + M_{n+1}{(\hat{k})}_{iu}). \label{Q-update1}
\end{eqnarray}
In view of the fact $b(n) = o(a(n))$, the coupled iterates (\ref{Q-update1}), (\ref{lambda-update}) form a two time scale stochastic approximation algorithm in the sense of \cite{BorkarBook}, section 6.1, with (\ref{Q-update1}) operating on the faster time scale and (\ref{lambda-update}) on the slower time scale. We exploit this fact later in the convergence analysis.

\section{Numerical examples}

Let us illustrate the proposed scheme with two examples, both with statistically identical arms (i.e., the transition matrix and reward do not depend on the arm).

\subsection{Example with circulant dynamics}

We first test our scheme on the example from \cite{Moka}. The example has four states
and the dynamics are circulant: when an arm is passive ($u=0$), resp. active ($u=1$),
the state evolves according to the transition probability matrices
$$
P_0=\left[\begin{array}{cccc}
{\small 1/2} & 0 & 0 & 1/2\\
1/2 & 1/2 & 0 & 0\\
0 & 1/2 & 1/2 & 0\\
0 & 0 & 1/2 & 1/2
\end{array}\right], \quad \mbox{and} \quad
P_1=P_0^T,
$$
respectively. The rewards do not depend on the action and are given by
$r(1,0)=r(1,1)=-1$, $r(2,0)=r(2,1)=0$, $r(3,0)=r(3,1)=0$, and $r(4,0)=r(4,1)=1$.
Intuitively, there is a preference to activate an arm when the arm is in
state 3. Indeed, the exact values of the Whittle indices, calculated in \cite{Moka}, are
as follows: $\lambda(1)=-1/2$, $\lambda(2)=1/2$, $\lambda(3)=1$, and $\lambda(4)=-1$, which
give priority to state 3.
Consider a scenario with $N=100$ arms, out of which $M=20$ are active
at each time. We initialize our algorithm with $\lambda_0(i)=0$,
and $Q(i,u)=r(i,u)$, $\forall i \in S$.

In this example, we assumed the shared memory architecture and took full advantage
of the fact that the arms are statistically identical. This helps to collect the statistics very
quickly and results in a rapid convergence of the algorithm.
We first set the exploration parameter as $\epsilon=0.1$.

\begin{figure}[ht]
      \centering
      \includegraphics[scale=0.3]{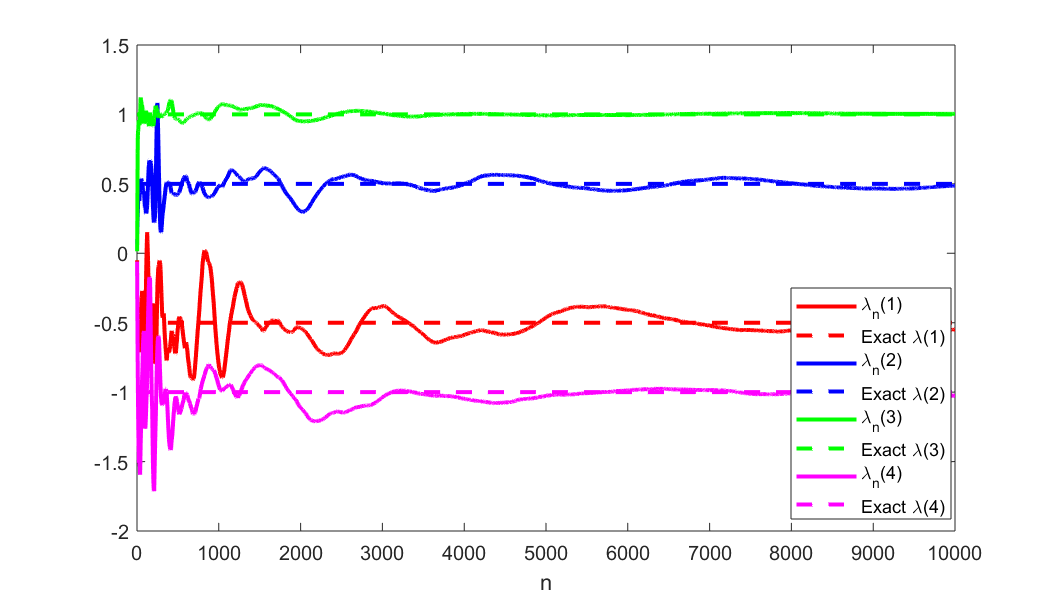}
      \caption{Estimated (solid lines) and exact (dash lines) Whittle indices in the example with circulant dynamics.}
      \label{fig:lambdasonpolicyud}
\end{figure}

\begin{figure}[ht]
      \centering
      \includegraphics[scale=0.3]{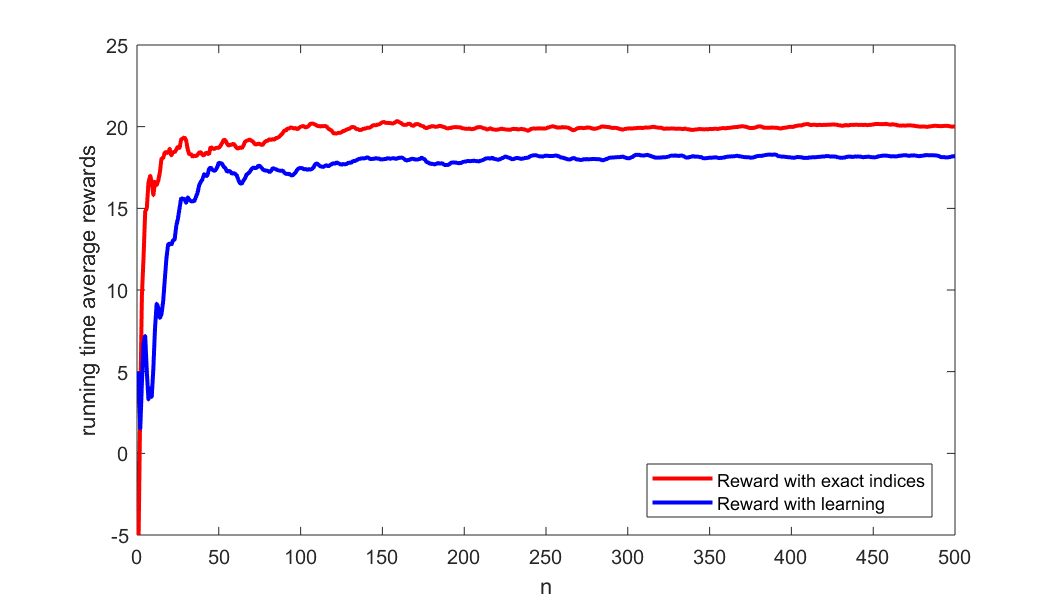}
      \caption{Rewards comparison in the circulant dynamics ($\epsilon=0.1$).}
      \label{fig:RewardsUDeps0_1}
\end{figure}

In Figure~\ref{fig:lambdasonpolicyud}
we present the convergence of the estimated values of the Whittle indices (see  (\ref{lambda-update}))
to the exact values. In Figure~\ref{fig:RewardsUDeps0_1}, we present the comparison
of the running average reward obtained by our algorithm with that of the algorithm
based on the use of the exact Whittle indices from the beginning. We see that the average
rewards stabilize in both approaches already after 250 iterations. The 10\% loss of efficiency
of our scheme with respect to the approach using the exact Whittle indices is due to the fact that
we spend 10\% of effort on pure exploration. This actually can be mitigated by decreasing the exploration
parameter with time. We notice that as predicted by the theory and confirmed by
Figure~\ref{fig:lambdasonpolicyud} the estimated Whittle indices in our algorithm converge
to the true values.

We also note that the convergence of the running average reward is significantly faster than
the convergence of the estimates of the Whittle indices (compare Figure~\ref{fig:RewardsUDeps0_1} vs
Figure~\ref{fig:lambdasonpolicyud}). This is because the control policy used depends only on the ordinal
comparison of the estimated Whittle indices and their order settles much faster than their actual numerical
values.

\begin{figure}[ht]
      \centering
      \includegraphics[scale=0.3]{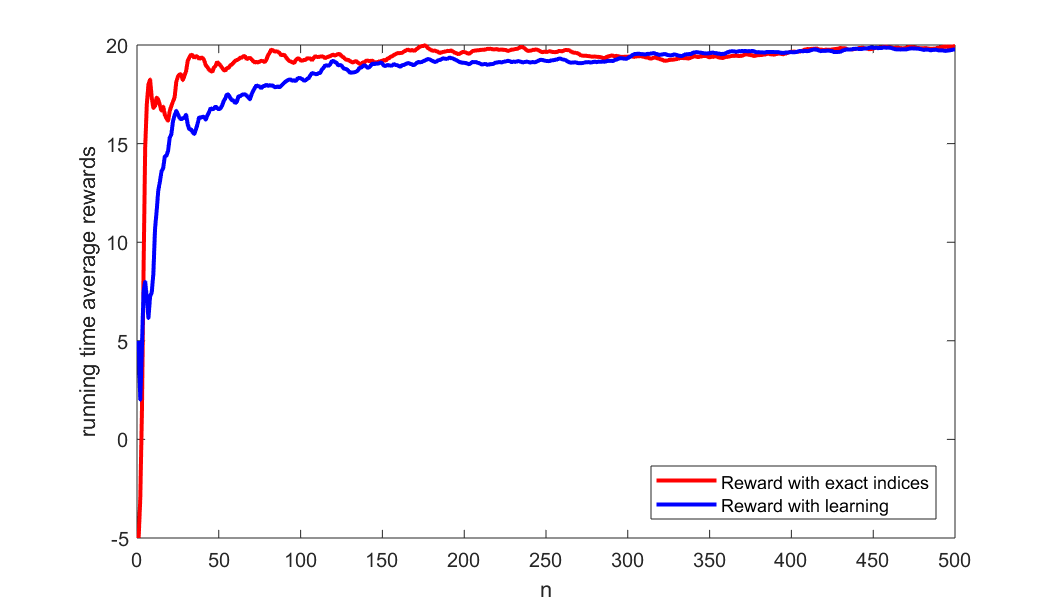}
      \caption{Rewards comparison in the circulant dynamics ($\epsilon=0.01$).}
      \label{fig:RewardsUDeps0_01}
\end{figure}

If we set the exploration parameter as $\epsilon=0.01$, there is hardly any loss of efficiency of our scheme
with respect to the scheme using the exact Whittle indices (see Figure~\ref{fig:RewardsUDeps0_01}).
Remarkably, the convergence of the running time averaged reward does not seem to suffer.
Of course, the convergence of the estimated Whittle indices to the exact values is now slower.
However, since the Whittle indices form a discrete set
with generous spacing, what matters is actually the ordinal ranking produced by the estimated Whittle
indices, which is quite robust, and not their proximity to the exact values.

This controlled chain in fact is not unichain, as under some stationary policies, it splits into two communicating classes. However, any state is reachable from any other under some control, as a result of which the optimal cost does not depend on the initial state and the dynamic programming equation (\ref{DP}) remains valid.

\subsection{Example with restart}

Now we consider an example where the active action forces an arm
to restart from some state. Specifically, we consider an example with 5 states,
where in the passive mode ($u=0$) an arm has tendency to go up the state space, i.e.,
$$
P_0=\left[\begin{array}{ccccc}
1/10 & 9/10 & 0 & 0 & 0\\
1/10 & 0 & 9/10 & 0 & 0\\
1/10 & 0 & 0 & 9/10 & 0\\
1/10 & 0 & 0 & 0 & 9/10\\
1/10 & 0 & 0 & 0 & 9/10
\end{array}\right],
$$
whereas in the active mode ($u=1$) the arm restarts from state 1 with probability 1, i.e.,
$$
P_1=\left[\begin{array}{ccccc}
1 & 0 & 0 & 0 & 0\\
1 & 0 & 0 & 0 & 0\\
1 & 0 & 0 & 0 & 0\\
1 & 0 & 0 & 0 & 0\\
1 & 0 & 0 & 0 & 0
\end{array}\right].
$$
The rewards in the passive mode are given by $r(k,0)=a^k$ (in our numerical experiments,
we have taken $a=0.9$) and the rewards in the active mode are all zero.

At least three facts have motivated us to choose this example. Bandits with restarting
dynamics have several applications such as congestion control \cite{Aetal13,ABSCDC},
web crawling \cite{ephemeral,Allerton,Nino14} and machine maintenance \cite{Glaze05}.
Their Whittle indices can be easily calculated, see e.g., \cite{Jacko,Larranaga}. The upper states
are much less visited, if at all, which poses a challenge for learning.

As in the previous example, we consider the scenario with $N=100$ arms out of which
$M=20$ are active at each time step. The exact Whittle indices are given by:
$\lambda(1)=-0.9$, $\lambda(2)=-0.73$, $\lambda(3)=-0.5$, $\lambda(4)=-0.26$,
and $\lambda(5)=-0.01$. We initialize the algorithm with $\lambda_0(i)=0$,
and $Q(i,u)=r(i,u)$, $\forall i \in S$.

\begin{figure}[ht]
      \centering
      \includegraphics[scale=0.3]{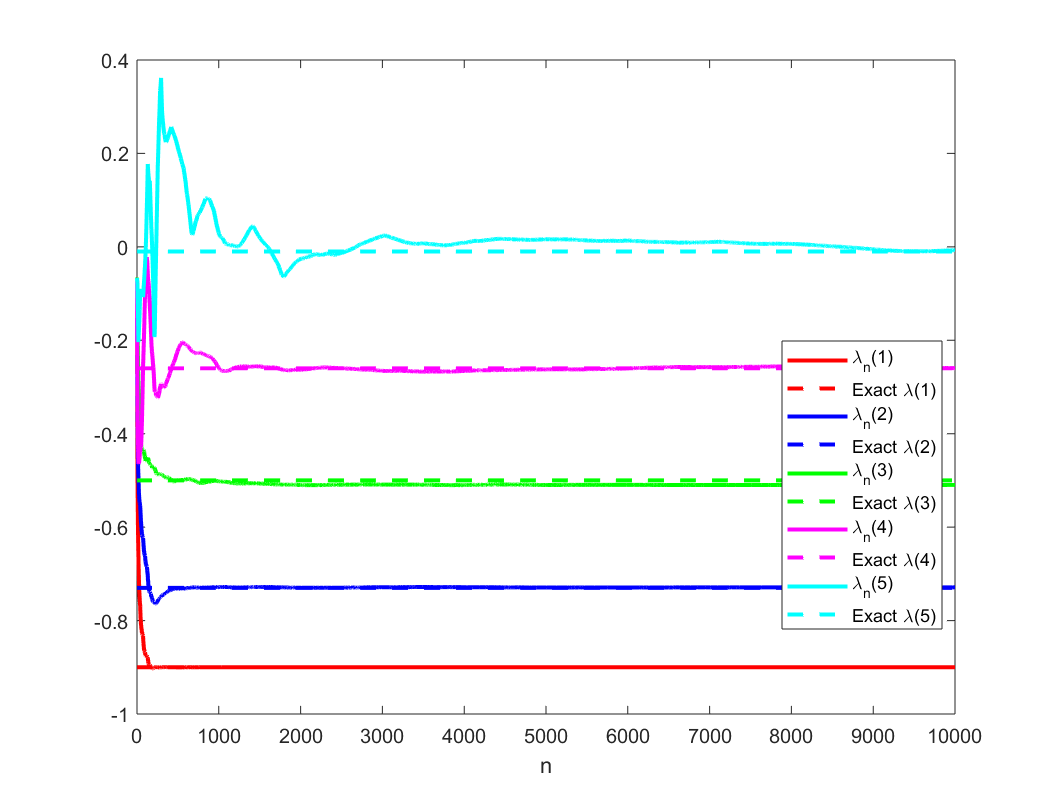}
      \caption{Estimated (solid lines) and exact (dash lines) Whittle indices in the example with restart.}
      \label{fig:lambdasonpolicyrestart}
\end{figure}

In Figure~\ref{fig:lambdasonpolicyrestart} we plot
the evolution of the estimated Whittle indices with $\epsilon=0.1$. As expected in this example,
the non-homogeneous structure of the state space poses some problems for learning
in comparison with the more symmetric example with circulant dynamics.
It takes noticeably longer time to learn the Whittle indices for the upper states 4 and 5
in comparison with the lower states 1, 2 and 3.

\begin{figure}[ht]
      \centering
      \includegraphics[scale=0.3]{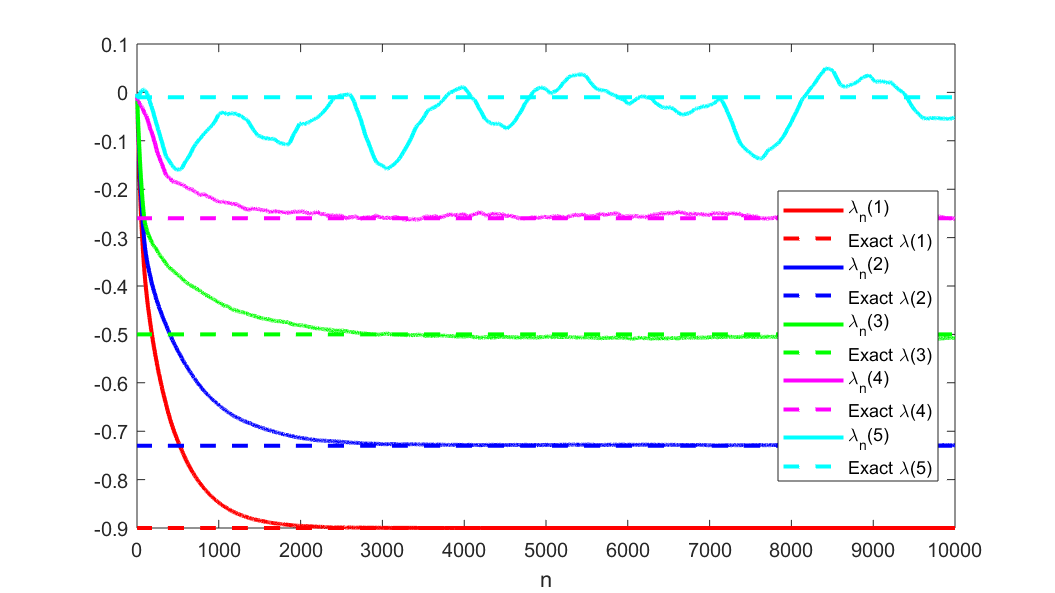}
      \caption{Estimated (solid lines) and exact (dash lines) Whittle indices in the example with restart.
      Constant step sizes: $a=0.02$, $b=0.005$.}
      \label{fig:lambdasonpolicyrestartconststep}
\end{figure}

So far, we have applied  decreasing  stepsizes recommended in (\ref{step_sizes}).
In practice one could also apply constant stepsizes.
For instance, in Figure~\ref{fig:lambdasonpolicyrestartconststep} we used constant stepsizes $a=0.02$, $b=0.005$.
The results are fairly good for all the states except the top state 5.
However, the top state is visited rarely and thus the value of its
Whittle index is not really relevant for good control of the system. One clear
practical advantage of the constant stepsize is the possibility of using this
variant for tracking a slowly varying environment.

As a final remark for this section, we would like to mention that the application
of both standard Q-learning and neural network based reinforcement learning
(e.g., DQN \cite{Mnih2015}) to the system as a whole
is simply not feasible, since in the operation $\max_v Q(i,v)$ we need to search through
the space of size $2^{100}$ during each learning step. We have also tried
Monte Carlo approach, where we sample a significant number of possible actions and
compared $Q$-values, or their approximations in the DQN algorithm, for those actions.
We could not observe any noticeable improvement in the empirical average reward
even after a very large number ($10^5$) of iterations.

\section{Convergence analysis}

In addition to \textbf{(C0)}, we make the following assumptions:\\

\begin{itemize}

\item \textbf{(C1)} The stepsizes $\{a(n)\}$  satisfy: $a(n+1) \leq a(n)$ for sufficiently large $n$, $\sum_na(n)^{1 + \upsilon} < \infty$ for some $\upsilon > 0$,  and,
 for $x \in (0, 1)$,
$$\sup_n\frac{a(\lfloor xn\rfloor)}{a(n)} < \infty,$$
$$ \sup_{y \in [x,1]}\left|\frac{\sum_{m=0}^{\lfloor yn\rfloor}a(m)}{\sum_{m=0}^na(m)} - 1\right| \to 0,$$
and for all $x > 0$ and
$$N(n,x) := \min\{m \geq n : \sum_{k=n}^ma(k) \geq x\},$$
the limit
$$\lim_{n\uparrow\infty}\frac{\sum_{k=\nu(i,u,n)}^{\nu(i,u,N(n,x))}a(k)}{\sum_{k=\nu(j,v,n)}^{\nu(j,v,N(n,x))}a(k)}$$
exists a.s., for $i, j \in S, \ u, v \in \U $.

These are satisfied, e.g., by $a(n) = \frac{1}{n}$ or $\frac{1}{n\log n}$ from some $n$ on.\\

\item \textbf{(C2)} The problem is Whittle indexable.

\end{itemize}

\bigskip

We prove convergence of the above scheme to the desired limit using a combination of  results from the theory of stochastic approximation, in conjunction with \cite{Abounadi}. We  call the iteration (\ref{Q-update0}) \textit{synchronous} if  all  components of $Q_n^\alpha(\hat{k})$ are updated at the same time, i.e., the indicator $I\{X_n = i, U_n = u\}$
in (\ref{Q-update0})  is dropped and $\nu(i,u,n)= n \ \forall i, u$. Also, for updating the $(i,u)$-th component, $X_{n+1}$ is replaced by $\mathcal{X}_{n+1}(i,u)$, a simulated random variable independent of all else, with law $p( \cdot | i,u)$.  The iterate becomes
\begin{eqnarray}
\lefteqn{Q^\alpha_{n+1}(i,u;\hat{k}) = Q^\alpha_n(i,u;\hat{k}) \ + } \nonumber \\
\ \ \ && a(n)\Big((1 - u)(r(i,0) + \lambda_n(\hat{k}))  + \ ur(i,1) +  \nonumber \\
\ \ \ && \max_{v\in\{0,1\}}Q^\alpha_n(\mathcal{X}_{n+1}(i,u),v;\hat{k}) - \nonumber\\
\ \ \ && f(Q^\alpha_n(\hat{k})) -  Q^\alpha_n(i,u;\hat{k})\Big).
\label{synch}
\end{eqnarray}
This is legitimate only for off-line and therefore off-policy learning. It does, however, provide a step towards analyzing the fully \textit{asynchronous} update (\ref{Q-update0}) based on a single run $\{(X_n, U_n)\}$, which  updates only the $(X_n, U_n)$th component at the $n$-th time step\footnote{One can  consider more general forms of asynchrony where some but not all, and not necessarily only one, components are updated at each time. The analysis will be similar.}.

Our analysis of the coupled iterations (\ref{Q-update0})-(\ref{lambda-update}) uses the two time scale analysis of \cite{BorkarBook}, Section 6.1. To facilitate this, we first need the well-posedness of the limiting o.d.e.s. Lemma 1 paves the way for it. We also need a.s.\ boundedness of the iterates. Lemmas 2 and 3 establish this using the criterion of \cite{Bhat}. This requires verification of the assumptions of \cite{Bhat}, which is carried out in Lemma 2 for the synchronous case first for ease of exposition. Indeed, \cite{Bhat} deals with synchronous iterates. Lemma 3 provides the link to extend this to asynchronous iterates where our choice of stepsizes plays a key role. Theorem~1 then provides the convergence argument using the methodology of Section 6.1 of \cite{BorkarBook} combined with \cite{Karmakar}.\\

Recall the function $h$ defined in (\ref{Hech}). The limiting  o.d.e.\  for (\ref{Q-update}) with $\lambda_n$ frozen at $\lambda$ is ((3.4) in \cite{Abounadi})
\begin{equation}
\dot{Q}_t = h(Q_t, \lambda).  \label{ode1}
\end{equation}
This has as its globally asymptotically stable equilibrium  the solution $Q^*_{\lambda} = [[Q^*_{\lambda}(i,u)]]$ of (\ref{Q-DP}) with $f(Q^*_\lambda) = \beta_\lambda \ (:= \beta$ with its $\lambda$-dependence made explicit), see Theorem 3.4 of \cite{Abounadi}.   Define $g: \R^{|S|} \mapsto \R^{|S|}$ by:
$$g_i(\lambda) = Q^*_\lambda(i, 1) - Q^*_\lambda(i,0).$$

\noindent \textbf{Lemma 1} The map $\lambda \mapsto Q^*_\lambda$ is Lipschitz.\\

\noindent \textbf{Proof} We have $f(\hat{Q}^*_\lambda) = \hat{\beta}_\lambda := \lambda\times$ the stationary probability of the set of passive states $+$ the stationary expectation of $r(X_n, U_n)$, under the optimal policy. For any stationary policy $\varphi$, the reward for this problem would be likewise, i.e., affine in $\lambda$ with slope $\in [0, 1]$ and the constant offset bounded uniformly in $\varphi$.  Since $\hat{\beta}_\lambda$ is the maximum thereof over all policies for each $\lambda$,  $\lambda \mapsto \hat{\beta}_\lambda$ is Lipschitz with Lipschitz constant $\leq 1$. Fix $i_0 \in S, u_0 \in \U$ and let $\tau :=$ the first time $X_n = i_0, U_n = u_0$. Then for a fixed stationary policy $\varphi : S \mapsto \U$, letting $Q^{(\varphi)}_\lambda$ denote the corresponding vector of Q-values, we have the representation (see, e.g., the arguments of Lemma 2.5, pp.\ 79-80, \cite{BorkarMC})
\begin{eqnarray*}
Q^{(\varphi)}_\lambda(i,u) &=& (1 - u)(\lambda + r(i, 0)) +ur(i, 1) +  \hat{\beta}_\lambda + \\
&&\sum_jp(j|i, \varphi(i))E_j\Bigg[\sum_{n=0}^{\tau - 1}((1 - \varphi(X_n))\times \\
&&(\lambda + r(X_n, 0)) + \varphi(X_n)r(X_n, 1)- \beta_\lambda)\Bigg].
\end{eqnarray*}
Using the fact that $E_j[\tau]$ is bounded uniformly in $j, \varphi$ under our hypotheses, it follows that the above is Lipschitz in $\lambda$ with a Lipschitz constant independent of $\varphi$. Hence $Q^*_\lambda(i,u) := \sup_\varphi Q^{(\varphi)}_\lambda(i,u)$ (see \textit{ibid.}) is Lipschitz. \hfill $\Box$

\noindent \textbf{Lemma 2} Under the hypotheses (C0), (C1) and (C2),
the updates of (\ref{lambda-update}),(\ref{synch}) remain a.s.\ bounded.\\


\noindent \textbf{Proof} This essentially follows from the results of \cite{Bhat}. Let us verify the assumptions (A1)-(A5)  of \cite{Bhat}, pp.\ 109-110,  one by one.

1.\ (A1) of \cite{Bhat} requires that $h, g$ are Lipschitz. This is obvious for $h$ and follows from Lemma 1 for $g$. \\

2.\ In the notation of \cite{Bhat}, $M^{(1)}_n, M_n^{(2)}$ correspond to
resp., $M_n{(\hat{k})}$ in (\ref{emm})
and the  process that is identically zero. Both of these are martingale difference sequences (the latter trivially so). Furthermore, $\forall n$,
    $$E\left[\|M_{n+1}(\hat{k})\|^2 | \F_n\right] \leq K\left(1 + \|Q_n(\hat{k})\|^2 + \|\lambda_n(\hat{k})\|^2\right)$$
 a.s.\   by the Lipschitz property of the functions involved. The zero process trivially satisfies such an inequality. This is precisely (A2) of \cite{Bhat}. \\

3.\ (A3) of \cite{Bhat}  requires that $\sum_na(n) = \sum_nb(n) = \infty$, $\sum_m(a(n)^2 + b(n)^2) < \infty$, and $b(n) = o(a(n))$, which hold here by assumption. \\

4.\ For (A4), consider the o.d.e.\  tracked by the iterates (\ref{Q-update0}) in the synchronous case. We need to use the analysis of \cite{Abounadi} for the \textit{synchronous} case for which Assumptions 2.1 and 2.2 of \textit{ibid.} suffice,  Assumptions 2.3 and 2.4 are not needed. Of these, the first is simply our assumption (C0), whereas the second is satisfied by the function $f$ defined in (\ref{eff}) by construction.
The limiting  o.d.e.\  with $\lambda_n$ frozen at $\lambda$ is (\ref{ode1}) has the
globally asymptotically stable equilibrium  $Q^*_{\lambda} = [[Q^*_{\lambda}(i,u)]]$.
(Here and until the end of the lemma proof we omit the notation for $\hat{k}$ from $Q^*$ and $\lambda$
to make the equations more transparent. We keep in mind that $Q^*$ and $\lambda$ depend on $\hat{k}$
throughout the proof.)
 The limit
$$h_{\infty}(Q, \lambda) := \lim_{c\uparrow\infty}\frac{h(cQ, c\lambda)}{c}$$
then corresponds to the Q-learning problem for average reward control with constant running reward $\equiv \lambda$ for passive states and zero reward for active states. As above, by Theorem 3.4 of \cite{Abounadi}, this converges to the unique $\hat{Q}^*_\lambda$ for which $f(\hat{Q}^*_\lambda) = \hat{\beta}_\lambda := \lambda\times$ the stationary probability of the set of passive states under the optimal policy.
By Lemma 2, $\hat{Q}^*_\lambda(i,u)$ is Lipschitz in $\lambda$.
Furthermore, for $\lambda = 0$, both the active and passive running rewards are zero, and therefore $\hat{\beta}_0 = 0$.
So the unique solution to (\ref{Q-DP}) with $f(\hat{Q}^*_0) = \hat{\beta}_0$ is the zero vector.
It follows that the o.d.e.\
\begin{equation}
\dot{Q}_t = h_\infty(Q_t, \lambda) \label{infode}
\end{equation}
has $\hat{Q}^*_\lambda$ as its unique asymptotically stable equilibrium, which reduces to the origin when $\lambda = 0$. This is precisely  (A4) of \cite{Bhat}. \\

5.\ Consider the limit
$$
g_\infty(\lambda) = \lim_{c\uparrow\infty}\frac{g(\hat{Q}^*_{c\lambda}, c\lambda)}{c}.
$$
Letting
$$\hat{r}_c(i,u) := \Big(ur(1,i) + (1 - u)(c\lambda + r(0,i))\Big)/c$$
denote the scaled running reward and $\beta^{(c)} := \beta/c$ the scaled optimal reward, both are seen to be uniformly bounded for $c \geq 1$.
 Divide both sides of equation (\ref{Q-DP})  by $c$ and let $c\uparrow\infty$.
For each $c \geq 1$, it becomes the counterpart of (\ref{Q-DP}) for running reward $r_c$ that remains uniformly bounded over $c \in [1, \infty)$.
Let $\tau$ be the first hitting time of a fixed state $i_0 \in S$ accessible from every other state as per (C0)  and $V_c$  the value function for the average reward problem with running reward $r_c$ and $V_c(i_0) = 0$. For now, we write Q-values as $Q^{(c)}(\cdot,\cdot)$ to show the $c$-dependence explicitly. Using a  standard representation for the value function (\cite{BorkarMC}, p.\ 79),
\begin{eqnarray*}
&&Q^{(c)}(i,u)/c \\
 &=& \hat{r}_c(i,u) - \beta^{(c)}  + \sum_jp(j|i,u)V_c(j) \\
&=& \hat{r}_c(i,u) - \beta^{(c)} + \sum_jp(j|i,u)\times \\
&&\max_{v \in SP}E\Big[\sum_{m=0}^\tau(r_c(X_m,v(X_m)) \\
&& - \ \beta^{(c)}) | X_0 = j \Big] \\
&\leq& C\left(1 + \max_{SP, \ j \in S}E\left[\tau | X_0 = j\right] \right)< \infty
\end{eqnarray*}
by (\ref{hittingtime}), for a suitable constant $C$. Thus $Q^{(c)}(\cdot, \cdot)/c, \beta^{(c)}$ remain bounded as $c\uparrow\infty$. Any limit point $(Q^{(\infty)}_\lambda(\cdot, \cdot),$ $\beta^{(\infty)}_\lambda)$ thereof (with the $\lambda$-dependence rendered explicit again) satisfies
\begin{eqnarray}
Q^{(\infty)}_\lambda(i,0) &=& \lambda - \beta^{(\infty)}_\lambda + \sum_jp(j|i,0)\times \nonumber \\
&& \ \ \ \ \ \ \max_v Q^{(\infty)}_\lambda(j,v),  \label{plus1} \\
Q^{(\infty)}_\lambda(i,1) &=& -\beta_\lambda^{(\infty)} +\sum_jp(j|i,1)\times \nonumber \\
&& \ \ \ \ \ \ \max_v Q^{(\infty)}_\lambda(j,v). \label{plus2}
\end{eqnarray}
Consider three distinct cases:\\

\begin{itemize}
\item \textit{Case 1:} For $\lambda > 0$ as  $c\uparrow\infty$, eventually $u = 0$ is  optimal for all states. Then $\beta_\lambda^{(\infty)} = \lambda$ and $Q^{(\infty)}(i,0) = \max_vQ^{(\infty)}(i,v) \ \forall i$. By (\ref{plus1}),
$$Q^{(\infty)}_\lambda(i,0) = \sum_jp(j|i,0)Q^{(\infty)}_\lambda(j,0) \ \forall i,$$
implying
$$Q^{(\infty)}_\lambda(i,0) \equiv \ \mbox{a constant} \ = \max_vQ^{(\infty)}(i, v)  \ \forall i.$$
Substracting  (\ref{plus1}) from (\ref{plus2}),
$$Q_\lambda^{(\infty)}(i,1) - Q_\lambda^{(\infty)}(i,0) = -\beta_\lambda^{(\infty)} = -\lambda.$$

\item \textit{Case 2:} For $\lambda < 0$
 as $c\uparrow \infty$, eventually $u = 1$ is optimal for all states. Equation (\ref{plus2})  then implies that $\beta_\lambda^{(\infty)} = 0$, otherwise the equation does not have a solution: Iterating (\ref{plus2}) leads to
$Q^{(\infty)}_\lambda(i,1) = -n\beta_\lambda^{(\infty)} +$ a bounded quantity. This becomes unbounded unless $\beta_\lambda^{(\infty)} = 0$. In turn, (\ref{plus2}) with $\beta_\lambda^{(\infty)} = 0$  leads to $Q_\lambda^{(\infty)}(i,1) \equiv$ a constant independent of $i$.  From (\ref{plus1}), we then have
$$Q^{(\infty)}_\lambda(i,1) - Q^{(\infty)}_\lambda(i,0) = -\lambda + \beta_\lambda^{(\infty)} = -\lambda.$$

\item \textit{Case 3:} For $\lambda = 0$, $\beta_\lambda^{(\infty)} = 0$ and the zero vector trivially satisfies (\ref{plus1}), (\ref{plus2}). The solution thereof is unique up to an additive scalar.
    This leads to
$$Q^{(\infty)}_\lambda(i,1)  - Q^{(\infty)}_\lambda(i,0) = 0 = -\lambda.$$
\end{itemize}
We have proved that $g_\infty(\lambda) = -\lambda \ \forall \lambda$. The limiting o.d.e.\
$\dot{\lambda}_t = g_\infty(\lambda_t) = -\lambda_t$
 has zero  as its unique globally asymptotically stable equilibrium. This verifies (A5) of \cite{Bhat}.

Theorem 10 $(iv)$ of \cite{Bhat}
then implies a.s.\ boundedness of the iterates, i.e.,
$$\sup_n|\lambda_n(\hat{k})| < \infty, \ \sup_n|Q_n^\alpha(i,u;\hat{k})| < \infty \ \forall \ i,\hat{k},u, \quad \mbox{a.s.}$$
\ \hfill $\Box$

\bigskip

\noindent \textbf{Lemma 3} Under the hypotheses (C0), (C1) and (C2), the updates of (\ref{Q-update0})-(\ref{lambda-update}) remain a.s.\ bounded.\\

\noindent \textbf{Proof} The only difference with Lemma 2 here is that the fast iterate (\ref{Q-update0}) is asynchronous. First we consider it in isolation, as in \cite{Bhat}, section 5.1. From the arguments leading to Theorem 8 therein, the crux is the behavior of the limiting o.d.e. From \cite{asyn} (see equation (2.11) therein),  an asynchronous variant tracks a limiting o.d.e.\ of the form
\begin{equation}
\dot{Q}_t = \Gamma_t h(Q_t, \lambda_t), \label{infode2}
\end{equation}
where $t \mapsto \Gamma_t$ is a diagonal matrix valued trajectory with non-negative entries on the diagonal of $\Gamma_t$ for each $t \geq 0$. (These reflect the relative frequencies with which different components get updated, see, e.g., \cite{BorkarBook}, Chapter 7.) Note that we have allowed time dependence of $\lambda$  so that the framework of section 4 of \cite{Bhat} becomes applicable. Our assumptions (C0) and (C1), together
with condition (\ref{frequent}) verify resp., Assumptions 2.1, 2.3, 2.4 of \cite{Abounadi}), whereas, as already observed, our function $f$ in (\ref{eff}) was explicitly chosen to satisfy Assumption 2.2 therein, viz., for $\1 :=$ the vector of all $1$'s, $f(\1) = 1$ and $f(x + c\1) = f(x) + c)$.
Thus as in \cite{asyn}, we have $\Gamma_t \equiv \frac{1}{2d}\times$ the identity matrix.
See, e.g.,  the concluding remark of section 3, p.\ 850, in \textit{ibid.}, where this claim follows from the proof of Theorem 3.2\footnote{See also the correction note of \textit{ibid.}} therein. This makes (\ref{infode2}) merely a time-scaled version of (\ref{ode1}). That (\ref{ode1}) has a unique asymptotically stable equilibrium is already established in Theorem 3.4, p.\ 689, of \cite{Abounadi}. Hence the arguments of \cite{Bhat} apply and the claim follows. \hfill $\Box$

This brings us to our main result.\\

\noindent \textbf{Theorem 1} Under the hypotheses (C0), (C1) and (C2),  $\lambda_n(\hat{k}) \to$ the Whittle index $\lambda(\hat{k})$ for all $\hat{k} \in S$, a.s.\\

\noindent \textbf{Proof} The claim follows from Theorem 2 of \cite{Karmakar}.  We begin by verifying the assumptions of \cite{Karmakar}.

1. The process $\{X_n\}$ takes values in a finite, hence compact state space, and the probability $P(X_{n+1} = i | \F_n)$ for $i \in S$ depends only on $X_n, U_n$ for each $n$. This verifies (A1) of \cite{Karmakar}. Furthermore, this dependence is trivially continuous because the latter take values in finite sets. This verifies (A5) of \textit{ibid.}

2.\ (A2), (A3) and (A4) of \cite{Karmakar} are  verified as follows: (A2) requires $h$  defined in (\ref{Hech}) to be Lipschitz, which it is. (A3)  requires $\{M_n(\hat{k})\}$ to be $\{\F_n\}$-martingales satisfying
$$E\left[\|M_{n+1}(\hat{k})\|^2 | \F_n\right] \leq K\left(1 +
\|Q_n(\hat{k})\|^2 + \|\lambda_n(\hat{k})\|^2\right).$$
This is also easily verified as in the beginning of the proof of Lemma 1 when we verified (A2) of \cite{Bhat}.  Finally, (A4) imposes the standard conditions on $\{a(n)\}, \{b(n)\}$, viz., $\sum_na(n) = \sum_nb(n) = \infty, \sum_n(a(n)^2 + b(n)^2) < \infty$ and $b(n) = o(a(n))$,  that we have already assumed\footnote{Note that in \cite{Karmakar}, $a(n) = o(b(n))$, so that the roles of $\{a(n)\}, \{b(n)\}$ are reversed.}.

3.\ Here we use the stronger condition (A6') of \cite{Karmakar} as opposed to (A6). For us, it reduces to the fact that the o.d.e.\ (\ref{infode}) has a unique globally asymptotically stable equilibrium $Q^*_\lambda :=$ the unique solution of (\ref{Q-DP}) corresponding to $f(Q^*_\lambda) = \beta_\lambda :=$ the optimal cost corresponding to $\lambda_n \equiv \lambda$ (Theorem 3.4, p.\ 689, \cite{Abounadi}). In the notation of \cite{Karmakar}, this corresponds to $\lambda(\theta)$.

4.\ (A7) assumes a.s.\ boundedness of the iterates, which we established in Lemmas 2 and 3.

The final observation we need is the fact that for each fixed $\hat{k}$, the slow iterates $\{\lambda_n(\hat{k})\}$ a.s.\ track the o.d.e.\
\begin{equation}
\dot{\Lambda}_t = Q^*_{\Lambda_t}(\hat{k}, 1) - Q^*_{\Lambda_t}(\hat{k}, 0). \label{lambdaode}
\end{equation}
This is a straightforward consequence of two timescale analysis of \cite{BorkarBook}, section 6.1, leading to Theorem 2, pp.\ 66-67, of \textit{ibid.}
If $\Lambda_t >$ the Whittle index $\lambda(\hat{k})$ of $\hat{k}$ (excess subsidy), the passive mode is preferred, i.e., $Q^*_{\Lambda_t}(\hat{k}, 0) > Q^*_{\Lambda_t}(\hat{k},1)$.
Then the r.h.s.\ is $< 0$ and $\Lambda_t$ decreases. Likewise, if the opposite (strict) inequality holds, the r.h.s.\ is $> 0$ and $\Lambda_t$ increases. Thus the trajectory $\Lambda_t, t \geq 0,$ remains bounded. Since any well-posed scalar o.d.e.\ with bounded trajectories must converge to an equilibrium,  $\Lambda_t$ converges to the $\Lambda$ satisfying $Q^*_{\Lambda}(\hat{k}, 1) =  Q^*_{\Lambda}(\hat{k}, 0)$, i.e.,  the Whittle index $\lambda(\hat{k})$.  This is unique by hypothesis. By theory of two time scale stochastic approximation (Theorem 2, section 6.1, \cite{BorkarBook}), we have $\lambda_n (\hat{k}) \to \lambda(\hat{k}), \ \forall \hat{k} \in S$, a.s.

 \hfill $\Box$

\section{Conclusions}

We have presented a novel Q-learning algorithm for Whittle indexable restless bandits and justified it both analytically and through numerical experiments.
The general philosophy  extends easily to related problems such as discounted rewards and related algorithms such as SARSA. An interesting future direction is to combine function approximation  with the present scheme  in order to handle large state spaces. Another open issue is to obtain convergence rates and regret bounds for this scheme. While some results are available for two time scale stochastic approximations in general (see, e.g., \cite{BorkarPattathil}), none seems available for two time scale algorithms with asynchronous iterates.

\bibliographystyle{plain}        
\bibliography{WhittleQlearningIni}

\end{document}